\documentclass{article}

\usepackage{PRIMEarxiv}

\errorcontextlines\maxdimen

\usepackage[utf8]{inputenc} 
\usepackage[T1]{fontenc}    
\usepackage{url}            
\usepackage{booktabs}       
\usepackage{nicefrac}       
\usepackage{microtype}      
\usepackage{fancyhdr}       
\usepackage{graphicx}
\usepackage{xcolor}

\usepackage{lineno,hyperref}
\hypersetup{colorlinks, citecolor=blue, linkcolor=red, urlcolor=green}
\usepackage{amsmath,amsfonts,amssymb}
\usepackage{graphics}
\usepackage{bm}
\usepackage{multirow}
\usepackage{algorithm}
\usepackage{algpseudocode}

\usepackage{mathtools}
\usepackage{siunitx}
\DeclarePairedDelimiterXPP\BigOSI[2]%
  {\mathcal{O}}{(}{)}{}%
  {\SI{#1}{#2}}

\title{Discovering stochastic partial differential equations from limited data using variational Bayes inference}

\author{Yogesh Chandrakant Mathpati\\
  Department of Applied Mechanics\\
  Indian Institute of Technology Delhi\\
  \texttt{ama212637@am.iitd.ac.in} \\
  \And
  Tapas Tripura\\
  Department of Applied Mechanics\\
  Indian Institute of Technology Delhi\\
  \texttt{tapas.t@am.iitd.ac.in} \\
  \And
  Rajdip Nayek\\
  Department of Applied Mechanics\\
  Indian Institute of Technology Delhi\\
  \texttt{rajdipn@am.iitd.ac.in} \\
  \And
  Souvik Chakraborty \\
  Department of Applied Mechanics\\
  Yardi School of Artificial Intelligence (ScAI)\\
  Indian Institute of Technology Delhi\\
  \texttt{souvik@am.iitd.ac.in} \\
}

\begin{document}
\maketitle

\begin{abstract}
We propose a novel framework for discovering Stochastic Partial Differential Equations (SPDEs) from data. The proposed approach combines the concepts of stochastic calculus, variational Bayes theory, and sparse learning. We propose the extended Kramers-Moyal expansion to express the drift and diffusion terms of an SPDE in terms of state responses and use Spike-and-Slab priors with sparse learning techniques to efficiently and accurately discover the underlying SPDEs. 
The proposed approach has been applied to three canonical SPDEs, (a) stochastic heat equation, (b) stochastic Allen-Cahn equation, and (c) stochastic Nagumo equation. Our results demonstrate that the proposed approach can accurately identify the underlying SPDEs with limited data.
This is the first attempt at discovering SPDEs from data, and it has significant implications for various scientific applications, such as climate modeling, financial forecasting, and chemical kinetics.
\end{abstract}


\section{Introduction}
Scientific machine learning has become an increasingly popular approach to solving complex problems in many scientific domains, including physics \cite{carleo2019machine}, chemistry \cite{von2020retrospective}, and engineering \cite{brunton2022data}. One important area of research in scientific machine learning is equation discovery, which aims to identify the underlying mathematical equations that govern a given physical system based on observational data. In recent years, a variety of methods have been developed for equation discovery, including sparse learning \cite{brunton2016discovering,nayek2021spike}, symbolic regression \cite{bongard2007automated,schmidt2009distilling}, and deep learning \cite{chen2021physics,both2021deepmod} based approaches. 
Work on discovering Lagrangian \cite{cranmer2020lagrangian,tripura2022learning} and Hamiltonian \cite{greydanus2019hamiltonian,han2021adaptable} can also be found in the literature. However, the majority of existing literature primarily focuses on the discovery of ordinary differential equations (ODEs) \cite{schaeffer2020extracting,fuentes2021equation} and partial differential equations (PDEs) \cite{rudy2017data,schaeffer2017learning} from data. Of late, researchers have started working on discovering stochastic differential equations (SDEs) from data \cite{boninsegna2018sparse,tripura2023sparse}. 
Nonetheless, there remains a significant gap in the field concerning the discovery of stochastic partial differential equations (SPDEs) from data, as no established framework or algorithm currently exists for this purpose. This paper aims to bridge this evident gap by presenting a novel algorithm for the Bayesian discovery of SPDEs from data. The objective is to propose a comprehensive framework that facilitates the discovery of the underlying SPDEs driving complex physical systems from observed data.

Early works in the model discovery of dynamical systems involve the use of the information criterion and expert knowledge \cite{nakamura2006comparative}. With an aim to automate the discovery, the works in \cite{bongard2007automated,schmidt2009distilling} later proposed a symbolic regression framework for identifying governing equations of dynamical systems from data. 
The utilization of compressive sensing in combination with nonlinear sparse optimization algorithms in the field of reduced-order modeling can be found in \cite{bright2013compressive,sargsyan2015nonlinear}.
With the rapid advancement in data-driven modeling, almost a decade later, since symbolic regression was introduced, Sparse Identification of Nonlinear Dynamics (SINDy) was proposed \cite{brunton2016discovering}. It offered a computationally cheaper alternative for identifying interpretable dynamical models from data.
Since then, the technique of sparse linear regression has been harnessed to great effect, yielding valuable insights and facilitating significant progress across a range of scientific disciplines, including sparse identification of chemical reactions in chemistry \cite {hoffmann2019reactive}, sparse identification of nonlinear dynamics for predictive control \cite{kaiser2018sparse}, sparse model selection
via an integral formulation \cite{schaeffer2017sparse}, recovering differential equations from short impulse response time-series data for model identification \cite{stender2019recovery}, sparse identification of biological network in \cite{mangan2016inferring}, model selection of dynamical system using sparse regression and information criteria \cite{mangan2017model}, system identification of structures with hysteresis \cite{lai2019sparse} and sparse learning of aerodynamics of bridges in structural systems \cite{li2019discovering}, identification of structured dynamical systems with limited data \cite{schaeffer2020extracting}, sparse modeling for state estimation in fluid mechanics \cite{loiseau2018sparse}, the discovery of partial differential equations \cite{rudy2017data,zhang2018robust}, etc. 

Deeply rooted in Bayesian statistics, the authors in \cite{nayek2021spike} proposed a sparse Bayesian approach to discover interpretable governing equations from data. Integrating concepts of Gibbs sampling with sparse learning through spike and slab prior, the authors were able to promote sparsity in the solution. In a simultaneous study in \cite{fuentes2021equation}, the authors used Relevance Vector Machine (RVM) to facilitate the equation discovery process. The use of these Bayesian approaches allows for model discovery in low data regimes and quantification of epistemic uncertainties in the identified model. While sparse learning with Gibbs sampler requires a high computational resource, the RVM-based approach often obtains non-parsimonious models, defeating the purpose of sparse learning. To address these shortcomings in a single framework, variational inference-based sparse learning with spike and slab prior is proposed in \cite{nayek2022equation}, which is recently extended to the discovery of partial differential equations by the authors \cite{more2023bayesian}.
In another recent approach, a symbolic genetic algorithm (SGA-PDE) is proposed for discovering PDEs from data \cite{chen2022symbolic}. 
Modern methods for distilling human interpretable governing equations also involve deep neural networks. Owing to the universal approximation theorem, deep neural networks can learn almost everything. Methodologies for data-driven equation discovery using deep learning can be found in \cite{bonneville2022bayesian,stephany2022pde,ren2022phycrnet,rao2022discovering}. 

All the frameworks proposed above pivot around the discovery of deterministic systems, i.e., systems that neglects the effect of random fluctuations arising due to natural agents. A more generalized approach is the data-driven discovery of stochastic forms of governing equations, e.g., stochastic ordinary and partial differential equations (SDEs and SPDEs) \cite{holden1996stochastic,lord2014introduction}. However, unlike the deterministic counterpart, the discovery of SDEs and SPDEs from data is an intriguing and challenging research area and yet to be explored to its full potential. 
For uncovering the stochastic forms, it is essential for the sparse learning approaches to accurately identify both the deterministic dynamics and the stochastic term enabling the exploration of the complex interplay between deterministic forces and random fluctuations. The stochastic SINDy proposed in \cite{boninsegna2018sparse} was the first approach aimed at identifying SDEs from data. Later the stochastic SINDy framework was extended to the discovery of SDEs with non-Gaussian L{\'e}vy type noise \cite{li2021data}. Sparse Bayesian learning (SBL) for systematic discovery of parsimonious forms of SDEs is proposed in \cite{wang2022data}. 
In recent work, a framework combining Kramers–Moyal expansion, sparsity–promoting spike and slab (SS) prior, and Gibbs-sampling for SDE discovery is proposed \cite{tripura2023sparse}. Later significant improvements have been applied to the framework to adapt it for real applications like model agnostic control \cite{tripura2022model} and predictive digital twins \cite{tripura2023probabilistic}. To overcome the computationally demanding formulation of the Gibbs sampling, an efficient variational Bayes alternative for SDE discovery from data is proposed in \cite{mathpati2023mantra}.

Despite steady progress in the SDE discovery, not a single work has been devoted to the discovery of SPDEs to date. In this paper, for the first time, we propose a novel variational Bayesian approach for discovering SPDEs from noisy and limited data. We propose the extended Kramers-Moyal expansion for expressing SPDEs in terms of observation data, which is the generalization of the Kramers-Moyal formula for SDEs. We combine it with sparse Bayesian learning using spike-and-slab (SS) prior for distilling a parsimonious PDE for the governing dynamical system. 
In particular, we exploit the extended Kramers-Moyal expansion to decouple the drift and diffusion terms of an SPDE. The decoupled drift and diffusion terms are expressed as a  weighted linear combination of candidate basis functions. A variational Bayesian algorithm then searches through the interpretable candidate functions to identify a simple and parsimonious model structure for the underlying physical system. Owing to the extended Kramers-Moyal expansion, the drift and diffusion terms can be discovered independently and in parallel, which further reduces the computational time.
The performance of the proposed method is demonstrated on three canonical SPDEs, including the 1D stochastic heat equation, 1D stochastic Allen-Cahn equation, and 1D stochastic Nagumo equation. The results obtained demonstrate the effectiveness of the proposed approach in accurately identifying the underlying SPDE.

The rest of the paper has been organized in the following manner: the problem statement is defined in section \ref{prob state}. In section \ref{sec:bayesian}, we preset the proposed variational Bayes approach for discovering SPDE from data. Results are shown in section \ref{problems}, using three different numerical problems. Section \ref{conclusion} finally provides discussion and concluding remarks.


\section{Problem statement}\label{prob state}
Let $\Omega \in \mathbb{R}^{n}$ be the domain of the solution space with boundary $\partial \Omega$. For all $x \in \Omega$, we consider the following general form of an SPDE,
\begin{equation}\label{eq:SPDE}
    \begin{aligned}
        &du(t,x) = f(u(t,x)) dt + g(u(t,x))dW(t,x), \; t>0, x\in \Omega, \\
        &u(t,0) = h(x), \; x\in \partial \Omega, \\
        &u(0,x) = u_{0}(x), x\in \Omega,
    \end{aligned}
\end{equation} 
where $u(t,x): \Omega \to \mathbb{R}$ is the solution of the SPDE, $x \in \mathbb{R}$ is spatial coordinate, $t>0$ is the temporal variables, $f(\cdot): \mathbb{R} \to \mathbb{R}$ is a nonlinear function that defines the dynamical evolution of the underlying process, and $g(\cdot): \mathbb{R} \to \mathbb{R}$ is a function that defines the noise strength.
The terms $f(\cdot)$ and $g(\cdot)$ are also known as drift and diffusion of the SPDE, respectively.
If $g(\cdot)$ is independent of the solution $u(t,x)$, the SPDE is said to have an additive noise, and in cases, where $g(u(t,x))$ is a function of $u(t,x)$, the SPDE is said to have multiplicative noise. Further, we denote $h(x)$ for $x\in \partial \Omega$ as the boundary condition, and $W(t,x) \in \mathbb{R}$ as the space-time Wiener process on $L^2(0,\partial \Omega)$. Following the standard terminologies in stochastic calculus, we refer to $f(\cdot)$ and $g(\cdot)$, respectively as the drift and diffusion of an SPDE throughout the manuscript.

To be able to work numerically, we discretize the spatial domain $\Omega = (0,a)$ into $N_x$ discrete coordinates. Further, we consider a temporal discretization of the time interval $t\in [0,T]$ for some $T>0$. The spatial discretization refers to the sensor locations at field condition and the temporal discretization represents the sampling frequency of the sensor. Representing $N_x$ and $N_t$ as the number of spatial and temporal points, and $N_s$ to be the number of ensembles of the measurements, we denote the collection of data as $\mathcal{D} = [ \mathbf{u} \in \mathbb{R}^{N_x \times N_t} ]_{j=1}^{N_s}$. 
Our aim is to identify the interpretable form of the given SPDE in Eq. \eqref{eq:SPDE}, however, since there is no direct approach we rediscover the SPDE by independently identifying the drift $f(u(t,x))$ and the diffusion $g(u(t,x))$ using the available data matrix $\mathcal{D}$.


\section{Variational Bayes framework for discovering SPDEs from limited data}\label{sec:bayesian} 
In this section, we propose a Bayesian framework for solving the problem discussed in section \ref{prob state}. 
While the discovery of deterministic systems focuses on uncovering deterministic relationships governing system dynamics, the discovery of SPDE goes a step further by distilling a mathematical form explaining the randomness of the governing PDEs. 
The proposed approach integrates the key concepts of stochastic calculus, variational Bayesian statistics, and sparse learning to learn the underlying SPDE from the time-history measurements $\mathcal{D}$. This section is divided into three parts. The first part extends the Kramers-Moyal formula for PDE in terms of data. In the second part, a sparse Bayesian framework is devised for inferring parsimonious models from data, and in the third part, a variational formulation for efficiently performing the sparse Bayesian learning is proposed.


\subsection{Extended Kramers-Moyal expansion for stochastic partial differential equations}
Consider the general form of an SPDE as follows,
\begin{equation}\label{eq:SPDE_kramers}
    du(t,x) = \left(Au(t,x) + \hat{F}(u(t,x)) \right)dt+ g(u(t,x))dW(t,x), \; t>0, x\in \Omega, u(0,x)=u_0(x),
\end{equation}
where $A: \mathbb{R} \to \mathbb{R}$ is the differential operator, $\hat{F}(\cdot)$ is the deterministic source term, $g(\cdot)$ and $dW(t,x)$ are, respectively the noise strength and Wiener process as defined in the previous section. 
noComparing with Eq. \eqref{eq:SPDE}, the drift can be identified as $f(u(t,x)) = Au(t,x) + \hat{F}(u(t,x))$. We then introduce an $J$-point spatial partition of the domain $\Omega=(0,a)$ using spatial increments $\Delta x$, where the solutions can be approximated using numerical schemes like finite difference as $[u(t,x_{1}),\ldots,u(t,x_{J})]$. If $\bm{u}_J(t)$ for $j=(J/\Delta x)$ is the finite difference approximation to the solutions for Eq. \eqref{eq:SPDE_kramers} then it is also said to satisfy the following equation,
\begin{equation}\label{eq:SDE_kramers}
   d\bm{u}_J(t) = \left({\mathbf{A}^D}{\bm{u}}_J(t) + \hat{\bm{F}}(\bm{u}_J(t))\right)dt + \bm{g}\left(\bm{u}_J(t)\right) d\bm{W}_{J}(t),
\end{equation}
where ${\mathbf{A}^D}$ represents the finite difference approximation to the differential operator $A$. Further, $\bm{u}_J(t) = [U_t(x_1),\ldots,U_t(x_j)]^T$ denotes the finite difference approximated solution at time $t$, and similarly $\bm{W}_J(t) = [W(t,x_1),\ldots,W(t,x_J)]^T$.
For illustration, one can consider the stochastic heat equation with additive Wiener noise,
\begin{equation}\label{eq:spdep2}
    du(t,x) = \left(\epsilon u_{xx}(t,x) + \hat{F}(u(t,x)) \right)dt + \sigma dW(t,x), \; \epsilon>0, t>0,x\in \Omega,u(0,x)=u_0(x).
\end{equation}
The central finite difference approximation of the second-order partial derivative in the drift term $f(u(t,x)) = \epsilon u_{xx}(t,x) + \hat{F}(u(t,x))$, yields the following stochastic differential equation (SDE),
\begin{equation}\label{spde:p1}
    \begin{aligned}
    {du(t,x_j}) = \left( \frac{\epsilon}{h^2} \left(u(t,x_{j+1})-2u(t,x_j)+u(t,x_{j-1})\right) + \hat{F}(u(t,x_j)) + r_j(t) \right)dt + \sigma dW(t,x).
    \end{aligned}
\end{equation}
Here $r_j(t)$ represents the $\mathcal{O}(\Delta x^2)$ residual error in finite difference approximation, which can be eliminated from the above expression using the limiting condition $\operatorname{lim}_{\Delta x \rightarrow 0} r_j(t) \to 0$.
Finally, with the homogeneous Dirichlet boundary conditions on $\Omega=(0,a)$ (i.e., $u(t,x_0) = u(t,x_{a}) = 0$) the systems of SDEs corresponding to stochastic heat equation can be derived as,
\begin{equation}
    {d\bm{u}(t,x}) = \left(\epsilon \mathbf{A}^D \bm{u}(t,x) + \hat{\bm{F}}(\bm{u}(t,x))\right)dt + \sigma d\bm{W}(t,x),
\end{equation}
where the finite difference operator ${\mathbf{A}^D}$ can be identified as,
\begin{equation}
    \mathbf{A}^D = \frac{1}{{\Delta x}^2}
    \left(\begin{array}{*{20}{c}}
        -2 & 1 & 0 & 0 & \ldots & 0 \\
        1 & -2 & 1 & 0 & \ldots & 0 \\
        0 & 1 & -2 & 1 & \ldots & 0\\[0.5em]
        \vdots &  &\ddots  & \ddots & & \vdots\\[0.5em]
        0  & \ldots & 0& 1 & -2 & 1\\
        0 & \ldots & 0 & 0 & 1 & -2\\
    \end{array}\right)
\end{equation}
Note that the approximated differential operator $\mathbf{A}^D$ can be adapted to other boundary conditions also, for e.g., Neumann and Robin boundary conditions. We return to the system of SDEs in Eq. \eqref{eq:SDE_kramers} and discretize using the Euler-Maruyama scheme. We denote the approximations of $u_J(t)$ by $U_{J,n}$ and $t_n = n\Delta t$, yielding,
\begin{equation}
    \Delta \bm{u}_{J,n} = \left({\mathbf{A}^D} {\bm{u}}_{J,n} + \hat{\bm{F}}(\bm{u}_{J,n})\right) \Delta t + \bm{g}(\bm{u}_{J,n}) \Delta \bm{W}_{J,n} + \mathcal{O}(\Delta t^2) + \mathcal{O}(\Delta \bm{W}), \; u_{J,0}=u_J(0),
\end{equation}
where $\Delta \bm{W}_{J,n} \triangleq \bm{W}_J(t_{n+1})-\bm{W}_J(t_n)$ is the Wiener increment, and $\bm{W}_{J}(t)\sim \mathcal{N}(\bm{0},t\mathbf{I}_{J})$. Here $\mathcal{O}(\Delta t^2)$ represents the first-order time approximation errors, and $\mathcal{O}(\Delta \bm{W})$ denotes the approximation error due to the absence of Wiener integrals of multiplicity greater than 1. The identification of the SPDEs is challenging due to the non-existence of time differentiability of the Wiener process. However, the drift and diffusion terms can be decoupled using the Kramers-Moyal expansion for SPDEs, which is yet to be derived. 
Let us consider, $p(u,x,t)$=${\rm{p}}(u,x,t|u_0,x_0,t_0)$ to be the transition probability density of the solution $u(t,x)$ of the SPDE in Eq. \eqref{eq:SDE_kramers}. The Kramers-Moyal expansion \cite{risken1996fokker} for the transition probability $p(u,x,t)$=${\rm{p}}(u,x,t|u_0,x_0,t_0)$ is written as,
\begin{equation}\label{eq:kramers_moyal}
    \dfrac{{\partial p(u,x,t)}}{{\partial t}} = {\sum\limits_{n = 1}^\infty {\left( { - \dfrac{\partial }{{\partial u}}} \right)} ^n}{M^{(n)}}(u,x,t)p(u,x,t),
\end{equation}
where the moments ${M^{(n)}}(\cdot)$ in the expansion are defined as,
\begin{equation}\label{eq:moments_kramers}
    {M^{(n)}}(u,x,t) = { \dfrac{1}{{n!}}\mathop {\lim }\limits_{{\Delta t}  \to 0} \dfrac{1}{{\Delta t} }\left\langle {{{\left| {u(t + {\Delta t},x) - u(t,x)} \right|}^n}} \right\rangle }.
\end{equation}
For $n=1$, ${M^{(1)}}(\cdot)$ yields the first Kramers-Moyal coefficient for the drift and for $n=2$, ${M^{(2)}}(\cdot)$ yields the second Kramers-Moyal coefficient for the diffusion. To find these coefficients, one will require to obtain the linear and quadratic variations of $\bm{u}(t,x)$ as,
\begin{equation}
    \begin{aligned}
        (\bm{u}_{J,n+1} - \bm{u}_{J,n}) =& \left( {\mathbf{A}^D} {\bm{u}}_{J,n} + \hat{\bm{F}} (\bm{u}_{J,n}) \right)\Delta t + \bm{g}(\bm{u}_{J,n}) \Delta \bm{W}_{J,n} + \mathcal{O}(\Delta t^2) + \mathcal{O}(\Delta \bm{W}^2), \\
        (\bm{u}_{J,n+1} - \bm{u}_{J,n})^2 =& \left( {\mathbf{A}^D} {\bm{u}}_{J,n} + \hat{\bm{F}} (\bm{u}_{J,n}) \right)^2 \Delta t^2 + \bm{g}(\bm{u}_{J,n})^2 \Delta \bm{W}_{J,n}^2 + \\
        &\left( {\mathbf{A}^D} {\bm{u}}_{J,n} + \hat{\bm{F}} (\bm{u}_{J,n}) \right) \bm{g}(\bm{u}_{J,n}) \Delta \bm{W}_{J,n} \Delta t + \mathcal{O}(\Delta t^2)  + \mathcal{O}(\Delta \bm{W}^2). \\
    \end{aligned}
\end{equation}
Upon taking expectations on both sides and utilizing the results $\mathbb{E}[\Delta {W}]=0$, $\mathbb{E}[(\Delta W)^{2}]=\Delta t$, and $\mathbb{E}[(\Delta W \Delta t)]=0$, from the It\^{o} calculus, the linear and quadratic variations of the solution $\bm{u}(J,n)$ are obtained as,
\begin{equation}
    \begin{aligned}
        \mathbb{E}\left[(\bm{u}_{J,n+1} - \bm{u}_{J,n})\right] &= \left( {\mathbf{A}^D} {\bm{u}}_{J,n} + \hat{\bm{F}} (\bm{u}_{J,n}) \right)\Delta t + \mathcal{O}(\Delta t^2), \\
        \mathbb{E}\left[(\bm{u}_{J,n+1} - \bm{u}_{J,n})^2\right] &= \left( {\mathbf{A}^D} {\bm{u}}_{J,n} + \hat{\bm{F}} (\bm{u}_{J,n}) \right)^2 \Delta t^2 + \bm{g}(\bm{u}_{J,n})^2 \Delta t + \mathcal{O}(\Delta t^2). \\
    \end{aligned}
\end{equation}
As $\Delta t \rightarrow 0$, the higher order time-dependent terms in the above expansion will vanish. Therefore, invoking limiting condition on time increment $\Delta t$, we obtain the coefficients of Kramers–Moyal expansion for SPDEs as follows,
\begin{equation}\label{eq:kramer1}
    \begin{aligned}
        \lim_{\Delta t \rightarrow 0} \frac{1}{\Delta t} \mathbb{E}\left[(\bm{u}_{J,n+1} - \bm{u}_{J,n})\right] &= \left( {\mathbf{A}^D} {\bm{u}}_{J,n} + \hat{\bm{F}} (\bm{u}_{J,n}) \right) = \bm{f}(\bm{u}_{J,n}), \\  
        \lim _{\Delta t \rightarrow 0} \frac{1}{\Delta t} \mathbb{E}\left[(\bm{u}_{J,n+1} - \bm{u}_{J,n})^2\right] &= \bm{g}(\bm{u}_{J,n})^2.
    \end{aligned}  
\end{equation}
The above expressions together provide the first two terms in the Kramers-Moyal expansion in Eq. \eqref{eq:kramers_moyal}. Therefore, in this section, we provide the evidence for existence of Kramers-Moyal expansion for SPDEs.
During the application of the proposed extended Kramers-Moyal expansion for discovering SPDEs from data, note that the diffusion component can be discovered only in quadratic form, as evident in the last expression of the above equation.


\subsection{Sparse Bayesian learning of stochastic partial differential equations}
In this section, we devise a Bayesian framework for identifying an interpretable form for the drift $\bm{f}(\bm{u}_{J,n})$ and diffusion $\bm{g}(\bm{u}_{J,n})$. We utilize the extended Kramers-Moyal formula (proposed in the previous section) to construct two separate and independent sparse linear regressions for the identification of drift and diffusion terms in parallel. 
To discover the drift and diffusion terms from data, we first express $\bm{f}(\bm{u}_{J,n})$ and $\bm{g}(\bm{u}_{J,n})^2$ as a weighted linear combination of certain basis functions as,
\begin{subequations}\label{eq:lib}
    \begin{align}
    {f}\left(\bm{u}(t,x)\right) & = \beta_{1}^{f} \ell_{1}^{f}\left(\bm{u}(t,x)\right)+\cdots+\beta_{k}^{f} \ell_{k}^{f}\left(\bm{u}(t,x)\right)+\cdots+\beta_{K}^{f} \ell_{K}^{f}\left(\bm{u}(t,x)\right), \\
    {g}\left(\bm{u}(t,x)\right)^2 & = \beta_{1}^{g} \ell_{1}^{g}\left(\bm{u}(t,x)\right)+\cdots+\beta_{k}^{g} \ell_{k}^{g}\left(\bm{u}(t,x)\right)+\cdots+\beta_{K}^{g} \ell_{K}^{g}\left(\bm{u}(t,x)\right).
    \end{align}
\end{subequations}
In the above equation, $\bm{\beta}_{k}^{f}$ and $\bm{\beta}_{k}^{g}$ denotes the weights of $k^{\text{th}}$ basis function for drift and diffusion components, respectively.
In general, the basis functions can have any mathematical form, for e.g., polynomials of $\bm{u}(t,x)$, partial derivatives of  $\bm{u}(t,x)$, trigonometric functions of $\bm{u}(t,x)$, element-wise multiplication between them, etc. 
In this work, the nature of the candidate functions that are considered inside the dictionary $\mathbf{D} \in \mathbb{R}^{N \times K}$ are illustrated below,
\begin{equation}\label{eq:dict}
    {\bf{D}} = \left[ \begin{array}{*{20}{c}}
    \vert & \vert & \vert &  & \vert & \vert &  & \vert & \vert &  \\
    {\bf{1}}&\bm{u}&\bm{u}\odot \bm{u}&\cdots&\bm{u}_x&\bm{u}_{xx}&\cdots&\bm{u}\odot \bm{u}_x&\bm{u}\odot \bm{u}_{xx} & \cdots \\
    \vert & \vert & \vert &  & \vert & \vert &  & \vert & \vert & 
    \end{array} \right],
\end{equation}
where $\bf{1}$ is the constant function of ones, $\bm{u}$ is the measurement matrix, $\bm{u}_{x}$ is first-order differentiation of $\bm{u}$ with respect to the subscript $x$, $\bm{u}_{xx}$ is second-order differentiation of $\bm{u}$ with respect to the subscript $x$, and $\odot$ is the Schur (element-wise) product.
We assume that we have a measurement matrix of size $\mathbf{u} \in \mathbb{R}^{N_x \times N_t\times N_s}$, the discrete measurements of $u(x,y,t)$ on $\Omega$, where $N_x$, $N_t$, and $N_t$ denote the number of spatial grid, time discretization, and ensemble size, respectively. To evaluate the dictionary and the coefficients of Kramers-Moyal expansion we perform a vectorization transformation $\bm{u}$ = vec($\mathbf{u}$), which reshapes the measurement matrix as, $\mathbf{u} \in \mathbb{R}^{N_x \times N_t\times N_s} \to \bm{u} \in \mathbb{R}^{N \times N_s}$ with $N \le N_xN_t$. The ensembles $N_s$ are crucial here and are required to perform the expectation in Eq. \eqref{eq:kramer1}. Also, for $N_s$ ensembles, one will get a collection of dictionaries $\hat{\mathbf{D}} \in \mathbb{R}^{N \times K \times N_s}$. Therefore, to perform the regression the expectation of the dictionary needs to be obtained with respect to the ensembles.
To proceed further we construct two separate regression problems using the Eq. \eqref{eq:kramer1} and \eqref{eq:lib}, which in compact form are given as follows,
\begin{subequations}\label{eq:compactspde0}
    \begin{align}
    \lim_{\Delta t \rightarrow 0} \frac{1}{\Delta t} \mathbb{E}\left[(\bm{u}_{J,n+1} - \bm{u}_{J,n})\right] & = f(\bm{u}(t,x)) + \bm{\varepsilon}^{f}
    = \mathbf{D}^{f}\bm{\beta}^{f}+\bm{\varepsilon}^{f}, \\
    \lim _{\Delta t \rightarrow 0} \frac{1}{\Delta t} \mathbb{E}\left[(\bm{u}_{J,n+1} - \bm{u}_{J,n})^2\right] & = g(\bm{u}(t,x)) + \bm{\varepsilon}^{g}
    = \mathbf{D}^{g}\bm{\beta}^{g}+\bm{\varepsilon}^{g},
    \end{align}
\end{subequations}
where $\bm{\varepsilon}^{f}$ and $\bm{\varepsilon}^{g}$ are the model mismatch error for drift and diffusion, respectively. 
Without loss of generality, we can represent the above regression problem using the following form,
\begin{equation}\label{eq:compactspde}
    \bm{Y}=\mathbf{D} \bm{\beta} + \bm{\varepsilon},
\end{equation}
where $\bm{Y} \in \mathbb{R}^{N}$ with $N \le N_x N_t$ denotes the $N$-dimensional target vector, $\mathbf{D} \in \mathbb{R}^{N \times K}$ with $K$ candidate basis functions denotes the design matrix (also a function of $\bm{u}(t,x)$), $\bm{\beta} \in \mathbb{R}^{K}$ is the weight vector, and $\bm{\varepsilon} \in \mathbb{R}^{N} \sim \mathcal{N}(0,1)$ is the residual error vector. Ideally, the weight vector $\bm{\beta}$ has non-zero entries for the basis functions that are present in the actual stochastic system and zero entries for other basis functions.
In case of drift identification the terms $\bm{Y}$, $\mathbf{D}$ and $\bm{\beta}$ will be replaced by $\bm{f}(\cdot)$, $\mathbf{D}^f$, and $\bm{\beta}^f$, whereas for diffusion identification these terms will be represented by 
$\bm{g}(\cdot)$, $\mathbf{D}^g$, and $\bm{\beta}^g$, respectively.
In Bayesian regression, to find the posterior distribution of the weight vector $\bm{\beta}$, one can employ the Bayes formula, where the likelihood function can be identified as,
\begin{equation}
    p(\bm{Y} \mid \bm{\beta}, \sigma^{2}) = \mathcal{N}\left(\mathbf{D} \bm{\beta}, \sigma^{2} \mathbf{I}_{N}\right),
\end{equation}
where $\mathbf{I}_{N} \in \mathbb{R}^{N \times N}$ is the identity matrix. To discover an interpretable expression for the underlying SPDE model, we invoke sparsity-promoting spike and slab (SS) priors in the Bayes formula. Our SS priors have a dirac-delta spike at zero and a flat Student's-t distribution spread over all possible values of weight parameters. The Dirac-delta spike shrinks all the small values of weight parameters to zero and allows a few weight parameters with significant values to escape the shrinkage.

In the SS prior, classification of the weight parameters into spike and slab distributions is achieved by using an indicator variable $Z_k$ for each weight component $\beta_k$. If a weight falls into the spike region, $Z_k$ takes a 0 value and if the weight falls into the slab region, $Z_k$ takes a value of 1 and only those terms will be retained in the final equation. Let $\bm{\beta}_r$ be a weight vector consisting of the weights for which the $Z_k$ takes a value 1. The SS prior is then defined as,
\begin{equation}
    p(\bm{\beta} \mid \bm{Z})=p_{s l a b}\left(\bm {\beta}_{r}\right) \prod_{k, Z_{k}=0} p_{s p i k e}\left(\beta_{k}\right),
\end{equation}
where $p_{spike}({\beta}_k) = \delta_0$ is the spike distribution, $p_{slab}(\bm{\beta}_r) = \mathcal{N}\left(\mathbf{0}, \sigma^{2} \vartheta_{s} \mathbf{I}_{r}\right)$ is the slab distribution, and $\vartheta_{s}$ is the slab variance. A hierarchical model is constructed by treating $\bm{Z}$ and $\sigma^{2}$ as random variables. The probability distributions are modelled as, $p(Z_{k} \mid p_{0}) =\operatorname{Bern}(p_{0})$ for $k=1 \ldots K$, and $p(\sigma^{2}) = \operatorname{IG}(\alpha_{\sigma}, \beta_{\sigma})$,
where the hyperparameters $\vartheta_{s}$, $p_0$, $\alpha_{\sigma}$, and $\beta_\sigma$ are taken as deterministic constants. 
The joint posterior distribution of the random variables $\bm{\beta}$, $\bm{Z}$, and $\sigma^2$ is given as,
\begin{equation} \label{eq:Bayes}
    \begin{aligned}
    p(\bm{\beta}, \bm{Z}, \sigma^{2} \mid \bm{Y}) &=\frac{p(\bm{Y} \mid \bm{\beta}, \sigma^{2}) p(\bm{\beta} \mid \bm{Z}, \sigma^{2}) p(\bm{Z}) p(\sigma^{2})}{p(\bm{Y})}, \\
    & \propto p(\bm{Y} \mid \bm{\beta}, \sigma^{2}) p(\bm{\beta} \mid \bm{Z}, \sigma^{2}) p(\bm{Z}) p(\sigma^{2}),
    \end{aligned}
\end{equation}
where $p(\bm{\beta}, \bm{Z}, \sigma^{2} \mid \bm{Y})$ is the joint posterior distribution of the random variables, $p(\bm{Y}|\bm{\beta}, \sigma^2)$ is the likelihood function, $p(\bm{\beta}\mid \bm{Z}, \sigma^2)$ is the prior distributions of the weight vector $\bm{\beta}$, $p(\bm{Z})$ is the the prior distributions of the latent variable vector $\bm{Z}$, $p(\sigma^2)$ is the prior distributions of the noise variance $\sigma^2$, and $p(\bm{Y})$ is the marginal likelihood. 
Due to the presence of the SS prior deriving an analytical expression for the posterior distribution is intractable. Thus one may use the Markov Chain Monte Carlo (MCMC) methods like the Metropolis–Hastings algorithm and Gibbs sampling methods; however, the MCMC methods become computationally expensive as the data point increases. To develop a computationally tractable framework, we propose to use the variational Bayes method which uses the KL divergence technique to  approximate the posterior distribution with the variational distribution. The procedure of iteratively updating the model is discussed in detail in the following section.


\subsection{Variational Bayesian inference for variable selection}
In this section, we propose an efficient variational Bayes method to approximate the joint posterior distribution $p\left(\bm{\beta}, \bm{Z}, \sigma^{2} \mid \bm{Y}\right)$ by simpler variational distributions. Firstly, due to the presence of a Dirac-delta function, it is necessary to reparameterize the linear regression model with SS prior to making it more suitable for the VB method. The parameterized SS prior is rewritten as,
\begin{equation}
    p(\bm{Y} \mid \bm{\beta}, \bm{Z}, \sigma^{2}) = \mathcal{N}\left(\mathbf{D} \bm{\Lambda} \bm{\beta}, \sigma^{2} \mathbf{I}_{N}\right),
\end{equation}
with the prior distributions,
\begin{equation}
    \begin{aligned}
    p(\sigma^{2}) = \operatorname{IG}\left(a_{\sigma}, b_{\sigma}\right), \;\;
    p(\beta_{k}) = \mathcal{N}\left(0, \sigma^{2} v_{s}\right), \;\;
    p(Z_{k}) = \operatorname{Bern}\left(p_{0}\right), i=1, \ldots, K,
    \end{aligned}
\end{equation}
where $\Lambda = \operatorname{diag}\left(Z_{1}, \ldots, Z_{K}\right)$. To approximate the true posterior distribution $p\left(\bm{\beta}, \bm{Z}, \sigma^{2} \mid \bm{Y}\right)$, let us consider some probability distribution $q(\bm{\beta}, \bm{Z}, \sigma^{2})$, which belongs to tractable family of variational distributions $Q$. 
To find the best approximate distribution $q^*\in Q$, the Kullback-Leibler divergence between the true posterior distribution $p\left(\bm{\beta}, \bm{Z}, \sigma^{2} \mid \bm{Y}\right)$ and the variational distribution $q(\bm{\beta}, \bm{Z}, \sigma^{2})$ is minimized as follows

\begin{align}
    q^{*}\left(\bm{\beta}, \bm{Z}, \sigma^{2}\right)
    &=\underset{q \in Q}{\arg \min } \operatorname{KL}\left[q\left(\bm{\beta}, \bm{Z}, \sigma^{2}\right) \| p\left(\bm{\beta}, \bm{Z}, \sigma^{2} \mid \bm{Y}\right)\right],   \\
    &=\underset{q \in Q}{\arg \min }\; \mathbb{E}_{q\left(\bm{\beta}, \bm{Z}, \sigma^{2}\right)}\left[ \ln \left({q\left(\bm{\beta}, \bm{Z}, \sigma^{2}\right)}\right) - \ln \left({p\left(\bm{\beta}, \bm{Z}, \sigma^{2} \mid \bm{Y}\right)}\right) \right] \label{eq:VB},
\end{align}
where $\mathbb{E}_{q\left(\bm{\beta}, \bm{Z}, \sigma^{2}\right)}[\cdot]$ is the expectation with respect to variational distribution $q\left(\bm{\beta}, \bm{Z}, \sigma^{2}\right)$. Using simple Bayes theorem and by making adjustments in Eq. \eqref{eq:VB}, we derive the evidence lower bound (ELBO) as follows,
\begin{equation}
    \begin{aligned}
    &\mathrm{KL}\left[q\left(\bm{\beta}, \bm{Z}, \sigma^{2}\right) \| p\left(\bm{\beta}, \bm{Z}, \sigma^{2} \mid \bm{Y}\right)\right] \\ 
    &={\operatorname{KL}\left[q\left(\bm{\beta}, \bm{Z}, \sigma^{2}\right) \| p\left(\bm{\beta}, \bm{Z}, \sigma^{2}\right)\right]-\mathbb{E}_{q\left(\bm{\beta}, \bm{Z}, \sigma^{2}\right)}\left[\ln p\left(\bm{Y} \mid \bm{\beta}, \bm{Z}, \sigma^{2}\right)\right]}+\ln p(\bm{Y}), \\
    &=\ln p(\bm{Y})-\mathrm{ELBO},
    \end{aligned}
\end{equation}
where $\text{ELBO} = \mathbb{E}_{q\left(\bm{\beta}, \bm{Z}, \sigma^{2}\right)}\left[\ln p\left(\bm{Y} \mid \bm{\beta}, \bm{Z}, \sigma^{2}\right)\right]-\mathrm{KL}\left[q\left(\bm{\beta}, \bm{Z}, \sigma^{2}\right) \| p\left(\bm{\beta}, \bm{Z}, \sigma^{2}\right)\right]$.
In the above equation, the term $\ln p(\bm{y})$ is  a constant with respect to the variational distribution $q\left(\bm{\beta}, \bm{Z}, \sigma^{2}\right)$ and the ELBO is considered as the lower bound of $\ln p(\bm{y})$. Therefore minimizing KL-divergence becomes equivalent to the maximization of the ELBO, yielding the following optimization problem,
\begin{equation}\label{eq:ELBO}
    q^{*}\left(\bm{\beta}, \bm{Z}, \sigma^{2}\right) = \underset{q \in Q}{\arg \max }\; \mathbb{E}_{q\left(\bm{\beta}, \bm{Z}, \sigma^{2}\right)}\left[\ln p\left(\bm{Y} \mid \bm{\beta}, \bm{Z}, \sigma^{2}\right)\right]-\mathrm{KL}\left[q\left(\bm{\beta}, \bm{Z}, \sigma^{2}\right) \| p\left(\bm{\beta}, \bm{Z}, \sigma^{2}\right)\right].
\end{equation}
For simplicity, we consider the random variables $\bm{\beta}$, $\bm{Z}$, and $\sigma^{2}$ to be independent, which helps in re-writing the variational distribution $q\left(\bm{\beta}, \bm{Z}, \sigma^{2}\right)$ in the following factorized form,
\begin{equation}
    q\left(\bm{\beta}, \bm{Z}, \sigma^{2}\right)=q(\bm{\beta}) q\left(\sigma^{2}\right) \prod_{i=1}^{K} q\left(Z_{i}\right). 
\end{equation}
The corresponding variational distributions of the variables $\bm{\theta}$, $\sigma^{2}$ and $\bm{Z}$ are given as, $q(\bm{\theta})=\mathcal{N}\left(\bm{\mu}^{q}, \bm{\Sigma}^{q}\right)$, $q\left(\sigma^{2}\right) =\mathcal{I} G\left(a_{\sigma}^{q}, b_{\sigma}^{q}\right)$, and $q\left(Z_{k}\right) = \operatorname{Bern} \left(w_{k}^{q} \right)$, for $i=1, \ldots, K$.
Here, $\bm{\mu}^{q}, \bm{\Sigma}^{q}, a_{\sigma}^{q}, b_{\sigma}^{q}, w_{i}^{q}$ are deterministic variational parameters, whose optimal values are obtained by maximizing the ELBO. The optimal parameters in order to maximize ELBO are found to satisfy the following relations,
\begin{align}
    q^{*}(\bm{\beta}) & \propto \mathbb{E}_{q(\bm{Z}) q\left(\sigma^{2}\right)}\left[\ln p\left(\bm{Y}, \bm{\beta}, \bm{Z}, \sigma^{2}\right)\right] \\
    q^{*}(\bm{Z}) & \propto \mathbb{E}_{q(\bm{\beta}) q\left(\sigma^{2}\right)}\left[\ln p\left(\bm{Y}, \bm{\beta}, \bm{Z}, \sigma^{2}\right)\right] \\
    q^{*}\left(\sigma^{2}\right) & \propto \mathbb{E}_{q(\bm{\beta}) q(\bm{Z})}\left[\ln p\left(\bm{Y}, \bm{\beta}, \bm{Z}, \sigma^{2}\right)\right]
\end{align}
One can solve the above equations and obtain the following update expressions for the variational parameters,
\begin{subequations}\label{eq:variationa_param}
    \begin{align}
        \bm{\Sigma}^{q} &=\left[\tau\left(\left(\mathbf{D}^{T} \mathbf{D}\right) \odot \bm{\Omega}+v_{s}^{-1} \mathbf{I}_{K}\right)\right]^{-1} \\
        \bm{\mu}^{q} &=\tau \bm{\Sigma}^{q} \mathbf{W}^{q} \mathbf{D}^{T} \bm{Y} \\
        a_{\sigma}^{q} &=a_{\sigma}+0.5 N+0.5 K \\
        b_{\sigma}^{q} &=b_{\sigma}+0.5\left[\bm{Y}^{T} \bm{Y}-2 \bm{Y}^{T} \mathbf{D W}^{q} \bm{\mu}^{q}+\operatorname{tr}\left\{\left(\left(\mathbf{D}^{T} \mathbf{D}\right) \odot \bm{\Omega}+v_{s}^{-1} \mathbf{I}_{K}\right)\left(\bm{\mu}^{q} \bm{\mu}^{q T}+\bm{\Sigma}^{q}\right)\right\}\right], \\
        \eta_{k} &=\operatorname{logit}\left(p_{0}\right)-0.5 \tau\left(\left(\mu_{k}^{q}\right)^{2}+\Sigma_{k, k}^{q}\right) \bm{h}_{k}^{T} \bm{h}_{k}+\tau \bm{h}_{k}^{T}\left[\bm{Y} \mu_{k}^{q}-\mathbf{D}_{-k} \mathbf{W}_{-k}^{q}\left(\bm{\mu}_{-k}^{q} \mu_{k}^{q}+\bm{\Sigma}_{-k, k}^{q}\right)\right] \\
        w_{k}^{q} &=\operatorname{expit}\left(\eta_{k}\right)
    \end{align}
\end{subequations}
where $\tau =a_{\sigma}^{q} / b_{\sigma}^{q}$,  $\operatorname{logit}(p_{0})=\ln(p_{0})-\ln(1-p_{0})$, $\operatorname{expit}(\eta_{k})= \exp(\eta_{k})/(1+\exp(\eta_{k}))$, $\bm{w}^{q}=[w_{1}^{q}, \ldots, w_{K}^{q}]^{T}$, $\mathbf{W}^{q}=\operatorname{diag}(\bm{w}^{q})$, and $\bm{\Omega}=\bm{w}^{q} \bm{w}^{q T}+\mathbf{W}^{q}(\mathbf{I}_{K}-\mathbf{W}^{q})$. Further, $\bm{h}_{i}$ denotes the $i^{\text{th}}$ column of $\mathbf{D}$, and $\mathbf{D}_{-i}$ represents the dictionary matrix after removing the $i^{\text{th}}$ column. The symbol $\odot$ denotes the Schur product between two matrices.
Due to inter-dependency between the update expressions of variational parameters, as observed in Eq. \eqref{eq:variationa_param}, an iterative updating method is used to optimize the variational parameters. The values of variational parameters are cyclically updated based on the values of the variational parameters in the most recent iteration.
As a check of convergence, the difference between ELBO in two successive VB iterations is tracked against a pre-defined error $\rho$ as,
\begin{equation}
    \operatorname{ELBO}^{(j)}-\operatorname{ELBO}^{(j-1)}<\rho.
\end{equation}
The updating is stopped when the difference between the ELBO value of two successive iterations is smaller than a pre-defined value $\rho$. The ELBO at every iteration $j$ is computed using the simplified expression given below,
\begin{equation}\label{eq:elbo}
    \begin{aligned}
        \mathrm{ELBO}^{(j)} = & \kappa+a_{\sigma} \ln \left(b_{\sigma}\right)-\ln \Gamma\left(a_{\sigma}\right)+\ln \Gamma\left(a_{\sigma}^{(j)}\right)
        -a_{\sigma}^{(j)} \ln \Gamma\left(b_{\sigma}^{(j)}\right)
        +0.5 \ln \left|\bm{\Sigma}^{(j)}\right| \\
        &+\sum_{i=1}^{K}\left[w_{i}^{(j)} \ln \left(\frac{p_{0}}{w_{i}^{(j)}}\right) +\left(1-w_{i}^{(j)}\right) \ln \left(\frac{1-p_{0}}{1-w_{i}^{(j)}}\right)\right].
    \end{aligned}
\end{equation}
In the above equation, $\kappa = 0.5K-0.5N \ln (2\pi)-0.5K \ln(v_{s})$ is a constant and $\Gamma(\cdot)$ is the Gamma function. The variables $a_{\sigma}^{(j)}$, $b_{\sigma}^{(j)}$, $\bm{\mu}^{(j)}$, $\bm{\Sigma}^{(j)}$, $\bm{w}^{(j)}$ denote the variational parameters at the $j^{\text{th}}$ iteration, and the variables $a_{\sigma}^{*}$, $b_{\sigma}^{*}$, $\bm{\mu}^{*}$, $\bm{\Sigma}^{*}$, $\bm{w}^{*}$ denote the final converged values.
\begin{algorithm}[t!]
    \caption{Variational Bayesian Inference for the discovery of SPDEs from data}\label{algvecvec}
    \begin{algorithmic}[1]
        \Require{Measurements: ${\mathbf{u}}(t,x) \in \mathbb{R}^{N_t \times N_x \times N_s}$ and hyperparameters: $ p, \vartheta_{s}, a_{\sigma}$, $b_{\sigma}$, error bound: $\rho$.}
        \State{Reshape the measurement matrix ${\mathbf{u}}(t,x) \in \mathbb{R}^{N_t \times N_x \times N_s} \to {\bm{u}}(t,x) \in \mathbb{R}^{N_tN_x \times N_s}$.}
        \State{Construct the target vectors.} \Comment{Eq. \eqref{eq:kramer1}}
        \State{Construct the dictionary ${\mathbf{D}^f}$ and ${\mathbf{D}^g}$ using the candidate basis functions.}\Comment{Eq. \eqref{eq:dict}}
        \State{Initialise $\bm{w}^{(0)}$ using stochastic SINDy and set the initial value of $\tau$.}
        \While {$\operatorname{ELBO}^{(j)}-\operatorname{ELBO}^{(j-1)}<\rho$}
        \State{Update the variational parameters $\bm{\Sigma}^{q}$, $\bm{\mu}^{q}$, $a_{\sigma}^{q}$, $b_{\sigma}^{q}$, $\tau,\eta_{i}$, and $w_{i}^{q}$.}\label{step1} \Comment{Eq.\eqref{eq:variationa_param}}
        \State {Compute the ELBO using the set of variational parameters for the iteration- $(j)$.}\label{stepn} \Comment{Eq.\eqref{eq:elbo}}
        \State{Repeat steps \ref{step1}$ \to $\ref{stepn}.}
        \EndWhile
        \State{Obtain the converged variational parameters, $a_{\sigma}^{*}$, $b_{\sigma}^{*}$, $\bm{\mu}^{*}$, $\bm{\Sigma}^{*}$, $\bm{w}^{*}$.}
        \State{Include basis function in the final model with $w_{i}^{*} > 0.5$ for $i=1,\ldots,K$.}
        \Ensure{Estimated mean ${\bm{\hat{\mu}}}_{\beta}$ and covariance ${\bf{\hat{\Sigma}}}_{\beta}$ of the weight vector $\beta \in \mathbb{R}^{K}$.}
    \end{algorithmic}
\end{algorithm}
The final model is selected based on marginal posterior inclusion probability (PIP), $p\left(Z_{i}=1 \mid \bm{Y}\right) \triangleq (1/K)\sum_{j=1}^{K}{Z_{i}^{j}}$ for $i = 1,\ldots,K$. In the proposed VB inference, the estimated $w_{i}^{(j)}$s represents the variational approximation to the posterior probability of $p\left(Z_{i}=1 \mid \bm{Y}\right)$. Therefore, after the convergence of the VB algorithm, only those basis functions with $w_{i}^{*}>0.5$ are included in the final estimated model. This means only those basis functions which have an occurrence probability greater than half are selected in the final model; however, one can choose a higher probability to derive simpler models.
We denote the estimated mean and covariance of the weight vector $\bm{\beta}$ by $\hat{\bm{\mu}}_{\bm{\beta}}$ and $\hat{\bm{\Sigma}}_{\bm{\beta}}$. The entries of $\hat{\bm{\mu}}_{\bm{\beta}}$ and $\hat{\bm{\Sigma}}_{\bm{\beta}}$ are initialized as zero, and only entries corresponding to selected basis functions are populated with corresponding values from $\bm{\mu}^{*}$ and $\bm{\Sigma}^{*}$. 
For simplicity, the complete procedure of discovering an SPDE from data is illustrated in brief in Algorithm \ref{algvecvec}.
%


\section{Numerical experiments}\label{problems}
In this proposed section, the efficacy of the algorithm is investigated using three numerical problems. For all the examples, one second of observation data is simulated. We have employed the well-known finite difference method and Euler-Maruyama scheme to generate the synthetic data.
The specific details about the number of realizations, sampling frequency, and time-step are provided separately with each example.
The objective is to discover the governing SPDEs from this data. The deterministic hyperparameters of the VB algorithm are adopted as $v_{s}=10$, $a_{\sigma}=10^{-4}$, $b_{\sigma}=10^{-4}$, and $p_{0}=0.1$. For initialization of the inclusion probability vector ${\bm{w}^{(0)}}$, the extended version of stochastic SINDy is utilized with a value of 0.3 of the sparsity constant. The initial value of $\tau$ is set to 1000. The ELBO threshold is set as $\rho= 10^{-6}$. To evaluate the accuracy of the proposed approach, we calculate the $L^2$ error and false positivity rate. 
Since this is the first attempt towards developing a framework for discovering SPDE from data, no benchmark methods exist. Therefore, we have created the following benchmark methods: (a) \textbf{Extended stochastic SINDy}: we formulated extended stochastic SINDy by combining the stochastic SINDy \cite{boninsegna2018sparse} with the extended Krammers Moyal formulation presented in this paper, (b) \textbf{Gibbs sampling based approach}: we formulated this by combining the extended Krammers Moyal based approach with Gibbs sampling based approach proposed in \cite{tripura2023sparse}. Owing to the fact that Gibbs sampler is extremely expensive, we have only used 100 particles in this work. With more particles, the results obtained using Gibbs sampling can be further improved; however, the computational cost will be exorbitant. 

\subsection{1D stochastic Allen-Cahn equation}\label{sec:ac}
As the first example, we consider the 1D stochastic Allen-Cahn equation, which is a type of stochastic partial differential equation (SPDE) frequently utilized in material science. This equation is particularly relevant to phase field models that describe the phase transition between two phases of a material, such as solid and liquid phases. The governing physics for stochastic Allen-Cahn equation takes the following form,
\begin{equation}\label{ac_eq}
    \begin{aligned}
        & du(t,x) = \left({\epsilon{u_{xx}(t,x)} + u(t,x) - u(t,x)^3 }\right)dt + \sigma dW(t,x), \; x \in [0,20], t\in [0,1], \\
        & u(0,x) = \left({ 1+exp(-(2-x)/\sqrt{2}) }\right)^{-1}, \; x \in [0,20], \\
        & u(t,x=20) = u(t,x=0), \; t\in [0,20],
    \end{aligned}
\end{equation} 
where $u(x,t)$ is the unknown function to be solved, $t>0$ is time, and $x \in \mathbb{R}$ is the coordinate in space. The parameter $\epsilon>0$ controls the rate of diffusion and the parameter $\sigma>0$ controls the strength of the noise.
The synthetic data is generated by solving the Eq. \eqref{ac_eq} within $t\in [0,1]$s with a time step of $\Delta{t} = 0.0025$s. The measurement space is discretized into $64$ discrete points using the finite difference method. After space discretization the Semi-implicit Euler–Maruyama is utilized to discretize in time using the values of parameters as $\epsilon=1$ and $\sigma=1$. To be able to accurately evaluate the expectation in extended Kramers-Moyal expansion, a total of $2000$ ensembles (i.e., $N_s= 2000$) of the measurement are generated.
\begin{figure}[ht!]
    \centering
    \includegraphics[width=\textwidth]{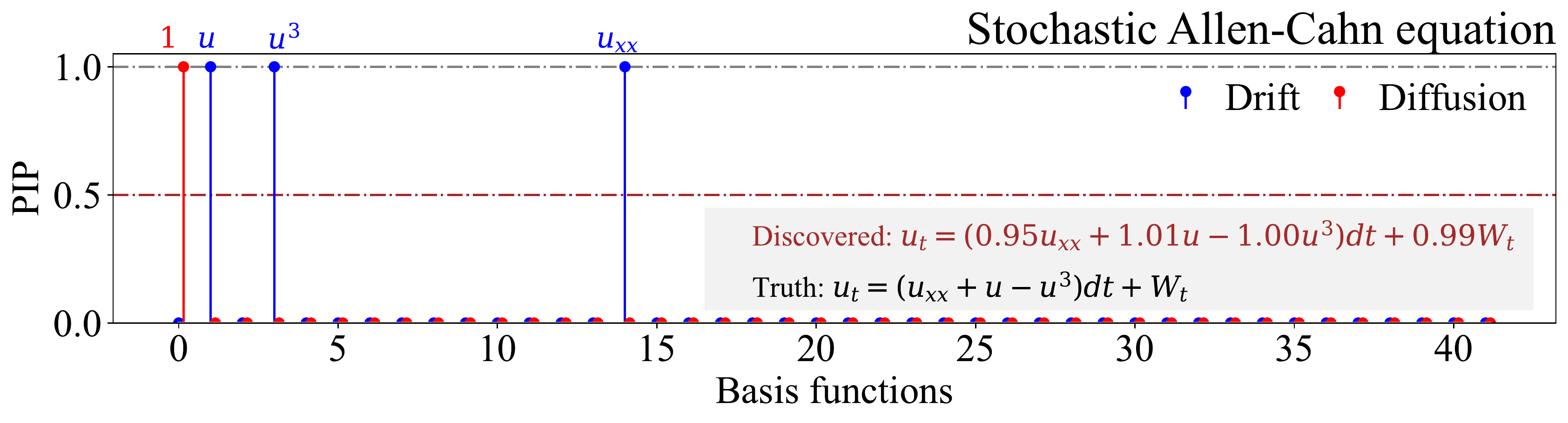}
    \caption{\textbf{Discovery of the 1D stochastic Allen Cahn PDE from data}. The regression problem contains 42 basis functions $\ell \in \mathbb{R}^{N}$ in the dictionary $\mathbf{D} \in \mathbb{R}^{N \times 42}$ and 42 sparse coefficients $\bm{\beta} \in \mathbb{R}^{42}$. Basis functions with PIP greater than a probability of 0.5 is selected in the final SPDE model. The set of basis functions present in the discovered drift model is $\{u, u^{3}, u_{xx}\}$, and for diffusion, the set contains $\{1\}$. The relative $L^2$ error in the discovered model is approximately 0.0259.}
    \label{fig:1d_ac}
\end{figure}

A total of $42$ basis functions are used in the dictionary $\textbf{D} \in \mathbb{R}^{N\times42}$, which in particular contains a vector of ones, polynomial functions of the measured response $u(t,x)$ upto order of 5, gradients of the measurement $u(t,x)$ with respect to the space $x$ upto order of $5$, and element-wise product between them. In Fig. \ref{fig:1d_ac}, the marginal PIP values of the basis functions in the drift and diffusion components for the Allen-Cahn PDE are shown. The discovered basis functions by the proposed approach are  $\left\{1, u, u^{3}, u_{xx}\right\}$. The true and discovered equations are as follows,
\begin{equation}
    \begin{array}{ll}
    \text{Ground Truth:} & du(t,x) = \left({u_{xx}(t,x) + u(t,x) - u(t,x)^3}\right) dt + dW(t,x),\\
    \text{Discovered:} & du(t,x) = (\underset{\pm 0.06}{0.95}u_{xx}(t,x) +\underset{\pm 0.05} {1.01}u(t,x) - \underset{\pm 0.05}{1.00} u(t,x)^3) dt +\underset{\pm 0.0002}{0.99} dW(t,x).  
    \end{array}
\end{equation}

The proposed variational Bayes accurately identifies all the relevant terms in the equation from the library and estimated parameters are also fairly accurate as compared to the actual values. It also gives the standard deviation of the parameters of the identified basis functions, which are indicative of the epistemic uncertainty associated with the discovered model. From identification of the basis functions to the near-accurate model parameters, the proposed variational Bayes framework utilizes only one second of measurements, sampled at a sampling frequency of $400$Hz.
In short, the identification results clearly indicate the robustness of the proposed framework in identifying the governing SPDE from limited measurements.
%
\begin{figure}[t!]
    \centering
    \includegraphics[width=\linewidth]{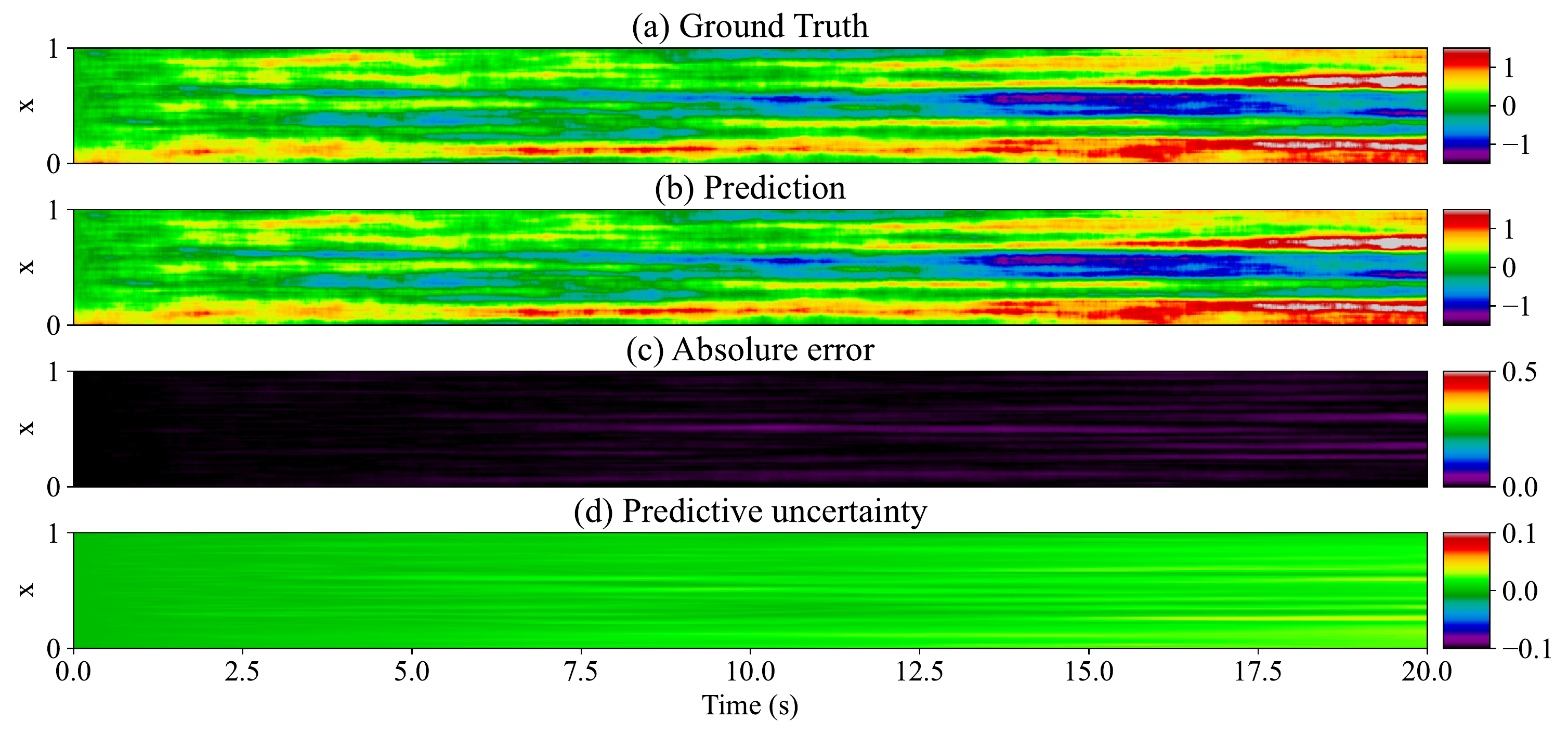}
    \caption{\textbf{Predictive performance and predictive uncertainty of the discovered 1D stochastic Allen-Cahn PDE against the Ground truth}. (a) The true solution of the actual SPDE. (b) The solution of the discovered stochastic Allen Cahn PDE. (c) The absolute error between the solutions of the true and identified system. (d) Predictive uncertainty in the discovered SPDE, obtained from 200 ensembles of the solution.}
    \label{fig:ac_all}
\end{figure}

The predictions for actual and identified SPDEs are shown in Fig. \ref{fig:ac_all}(a) and \ref{fig:ac_all}(b). The error between the predictions and predictive uncertainty is illustrated in Fig. \ref{fig:ac_all}(c) and \ref{fig:ac_all}(d). A total of 200 ensembles are generated to obtain the prediction results. From the prediction error plot, we observe that the discovered dynamical model is able to accurately learn the inherent dynamics of the actual SPDE model. 
Upon examining the standard deviation plots, it can be seen that the predictive uncertainty is increasing with time, which is expected and in line with other uncertainty quantification algorithms.
Further, the small value of both the standard deviation and prediction error indicates that the estimation provided by the proposed approach is consistent and reliable.

\subsection{1D stochastic Nagumo equation}
As the second example, we consider the stochastic Nagumo equation, which is a reduced model for wave propagation of the voltage in the axon of a neuron. The stochastic Nagumo equation is given as, 
\begin{equation}\label{nagumo_eq}
    \begin{aligned}
    & du(t,x) = \left(\epsilon u_{xx}(t,x) + u(t,x)(1-u(t,x))(u(t,x)-\alpha)\right) dt + \sigma dW(t,x), \; x \in [0,20], t \in [0,1], \\
    & u(0,x) = \left({1+exp(-(2-x)/\sqrt{2})}\right)^{-1}, \; x\in [0,20],\\
    & u(t,x=20) = u(t,x=0), \; t\in [0,1].
    \end{aligned}
\end{equation} 
In the above model, there are three model parameters, one is $\alpha \in \mathbb{R}$, which determines the speed of a wave traveling down the length of the axon, second is $\epsilon > 0$, which controls the rate of diffusion, and $\sigma>0$, which controls the strength of the noise. In data generation, $\alpha$, $\epsilon$, and $\sigma$ are taken as $-1/2$, $1$, and $1$, respectively.
To generate the synthetic data, we used the finite difference method for space discretization, and the semi-implicit Euler–Maruyama to discretize in time with a time step $\Delta t$. 
The temporal domain $t \in [0,1]$ is discretized using $\Delta t = 0.001$, and the spatial domain $x \in [0,20]$ is discretized into 64 discrete points. To apply the extended Kramers-Moyal expansion, a total of $2000$ ensembles (i.e., $N_s= 2000$) are generated.
\begin{figure}[ht!]
    \centering
    \includegraphics[width=\textwidth]{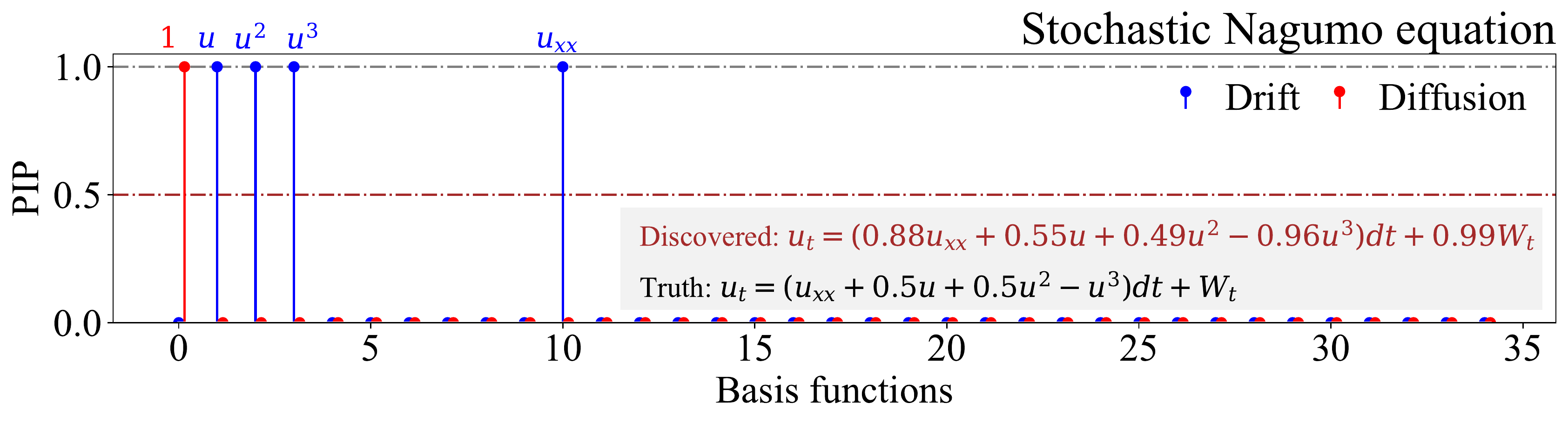}
    \caption{\textbf{Discovery of the 1D Stochastic Nagumo PDE from data}. The regression problem contains 35 basis functions $\ell \in \mathbb{R}^{N}$ in the dictionary $\mathbf{D} \in \mathbb{R}^{N \times 35}$ and 35 sparse coefficients $\bm{\beta} \in \mathbb{R}^{35}$. Basis functions with PIP greater than a probability of 0.5 are selected in the final SPDE model. The set of basis functions present in the discovered drift model is $\{u,u^2,u^3,u_{xx}\}$ and for diffusion the set contains $\{1\}$. The relative $L^2$ error in the discovered model is approximately 0.0731.}
    \label{fig:1d_nagumo}
\end{figure}

In this example, the dictionary $\mathbf{D} \in \mathbb{R}^{N\times 35}$ (consists of $35$ basis functions) contains similar library functions as the previous example, however, the orders of polynomials and partial derivatives are limited to 4. The marginal PIP values of the basis functions obtained by the proposed variational Bayes approach are shown in Fig. \ref{fig:1d_nagumo}. In the case of drift, the basis functions whose PIP values cross a probability of 0.5 are $\{u,u^2,u^3,u_{xx}\}$. Similarly, for the diffusion term, the only basis function is found to be the constant function ${1}$. The ground truth and discovered equation are illustrated below,
\begin{equation}
    \begin{array}{llllll}
    \text{Ground truth:} & du(t,x) = \left(u_{xx}(t,x) +0.5u(t,x) +0.5u(t,x)^2 -u(t,x)^3\right) dt + dW(t,{x}),\\
    \text{Discovered:} & du(t,x) = (\underset{\pm 0.07}{0.88}u_{xx}(t,x) +\underset{\pm 0.07} {0.55}u(t,x) +\underset{\pm 0.06}{0.49}u(t,x)^2 -\underset{\pm 0.09}{0.96} u^3)dt + \underset{\pm 0.0003}{0.99}dW(t,{x}).  
    \end{array}
\end{equation}
\begin{figure}[t!]
    \centering
    \includegraphics[width=\linewidth]{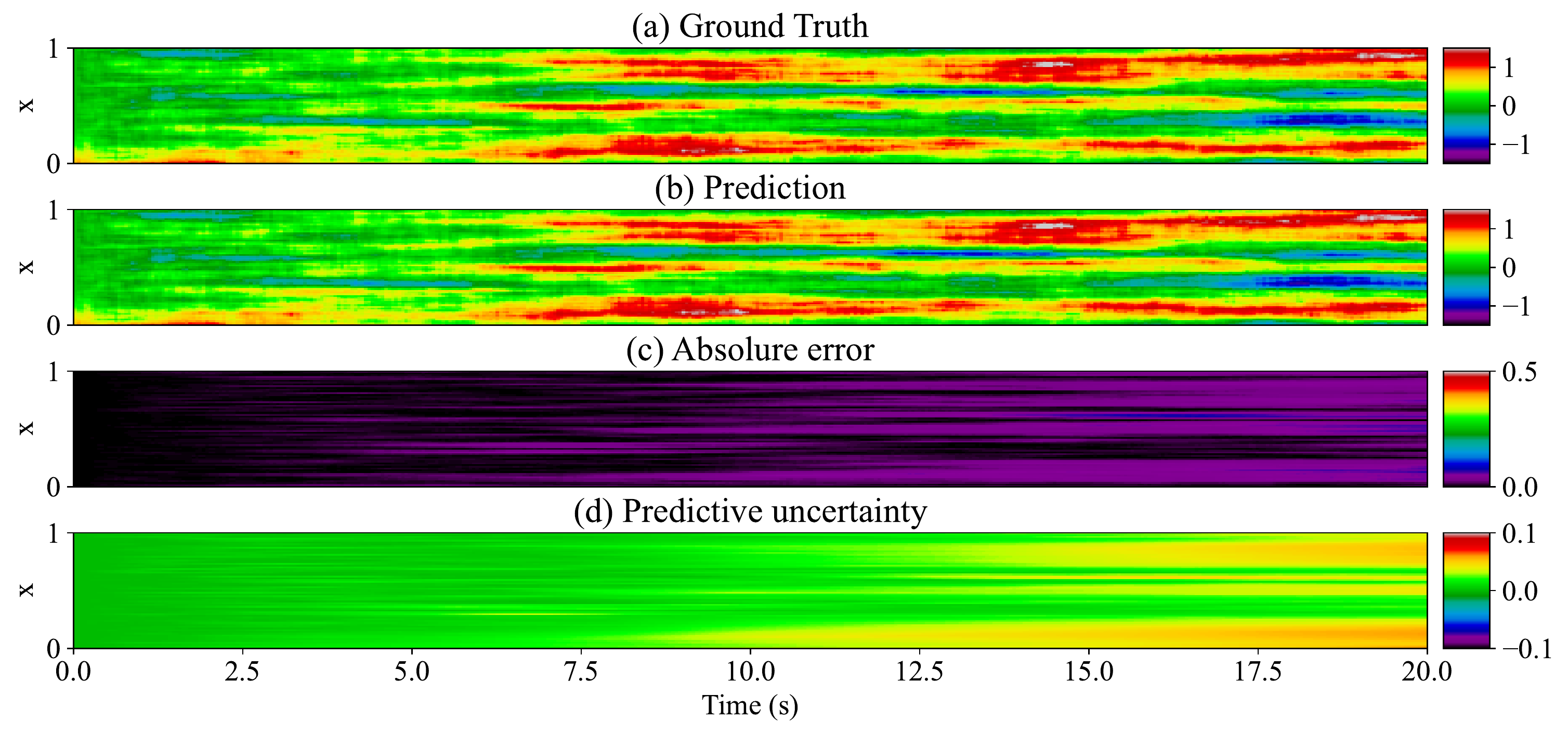}
    \caption{\textbf{Predictive performance and predictive uncertainty of the discovered 1D stochastic Nagumo PDE against the Ground truth}. (a) The true solution of the actual SPDE. (b) The solution of the discovered stochastic Nagumo PDE. (c) The absolute error between the solutions of the true and identified system. (d) Predictive uncertainty in the discovered SPDE, obtained from 200 ensembles of the solution.}
    \label{fig:nagumo_prediction}
\end{figure}
It is evident from these equations that the proposed VB framework is able to correctly identify all the terms in the true equation and the values of parameters associated with them are also near accurate.
The predictions obtained using the true and identified equations, the error between them, and the predictive uncertainty are shown in Fig. \ref{fig:nagumo_prediction}. The prediction error confirms that the discovered model accurately captures the inherent variability in the actual stochastic Nagumo equation.
Upon further examining these figures, it is apparent that the predictive uncertainty estimation provided by the proposed approach is increasing with time as in the previous example. From the results, therefore it can be conjectured that the predictions provided by the proposed VB framework are not only consistent and reliable with true solutions, but the discovered models can also capture the epistemic uncertainties.
%

\subsection{1D stochastic heat equation}\label{sec:heat}
As a last example, we consider the 1D stochastic heat equation. Unlike the deterministic counterpart, the stochastic heat equation considers the temperature fluctuations during the distribution of heat in a particular medium, over time. The stochastic heat equation is given as,
\begin{equation}
    \begin{aligned}
    & du(t,x) = \epsilon u_{xx}(t,x) dt + \sigma dW(t,x), \; x\in [0,20], t\in [0,1],\\
    & u(0,x) = \left(1+exp(-(2-x)/\sqrt{2})\right)^{-1}, \; x\in[0,20], \\
    & u(t,x=20) = u(t,x=0), \; t\in [0,1],
    \end{aligned}
\end{equation} 
where $\epsilon > 0$ is the thermal diffusivity of the medium, and $\sigma>0$ is the strength of the stochastic noise. 
For generating the synthetic data, the spatial domain $x\in [0,20]$ is discretized into $64$ discrete points using the finite difference scheme. Thereafter the semi-implicit Euler-Maruyama scheme is utilized with a time step of $\Delta{t} = 0.0025$s to obtain the measurements in $t\in [0,1]$s. A total of $2000$ ensembles (i.e., $N_s= 2000$) are generated to accurately evaluate the extended Kramers-Moyal expansion.
\begin{figure}[ht!]
    \centering
    \includegraphics[width=\textwidth]{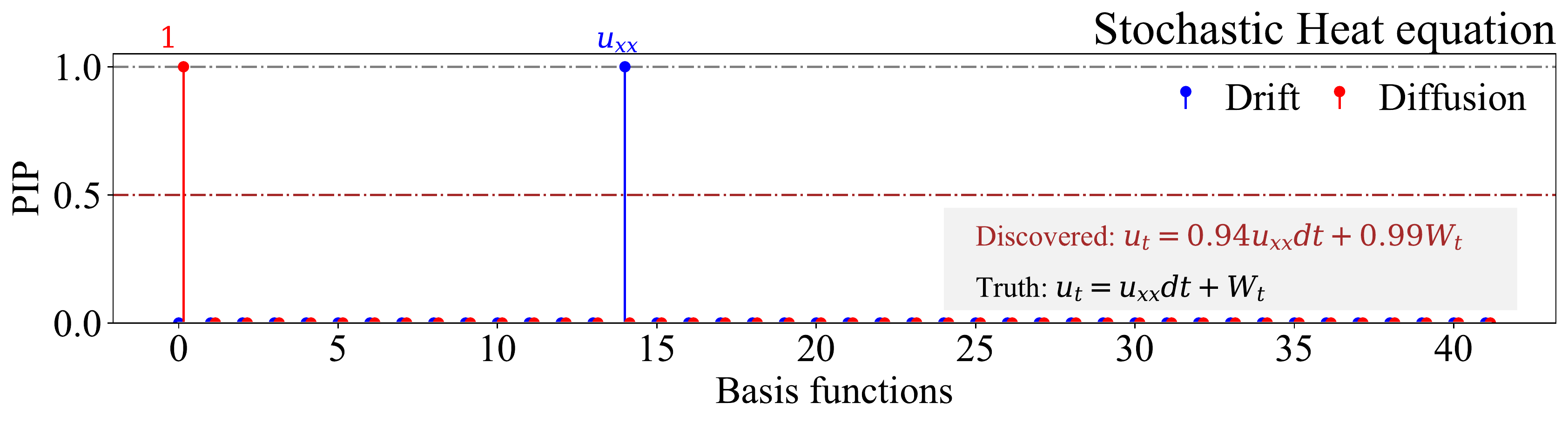}
    \caption{\textbf{Discovery of the 1D Stochastic Heat PDE from data}. The regression problem contains 42 basis functions $\ell \in \mathbb{R}^{N}$ in the dictionary $\mathbf{D} \in \mathbb{R}^{N \times 42}$ and 42 sparse coefficients $\bm{\beta} \in \mathbb{R}^{42}$. Basis functions with PIP greater than a probability of 0.5 are selected in the final SPDE model. The basis function present in the discovered drift model is $\{u_{xx}\}$ and for diffusion the basis function is $\{1\}$. The relative $L^2$ error in the discovered model is approximately 0.0430.}
    \label{fig:1d_heat}
\end{figure}
For the commencement of the VB algorithm, the dictionary $\mathbf{D} \in \mathbb{R}^{N \times 42}$ of 42 basis functions is created. The marginal PIP values of the basis functions in the Dictionary $\mathbf{D}$ are illustrated in Fig. \ref{fig:1d_heat}. It can be observed that the proposed VB algorithm identifies the correct basis functions of the stochastic heat equations, i.e., $1$ for diffusion, and $u_{xx}$ for drift. The ground truth and identified equations are as follows,
\begin{equation}
    \begin{array}{llllll}
    \text{Ground truth:} & du(t,x) = {u_{xx}(t,x)} dt + dW(t,{x}), \\
    \text{Discovered:} & du(t,x) = (\underset{\pm 0.05}{0.94}\,{u_{xx}(t,x)}) dt + \underset{\pm 0.003}{0.99} dW(t,{x}). 
    \end{array}
\end{equation}
It is evident from the above equations that the proposed VB algorithm identifies the terms in the equation correctly and the parameter values are also fairly accurate.

\begin{figure}[b!]
    \centering
    \includegraphics[width=\linewidth]{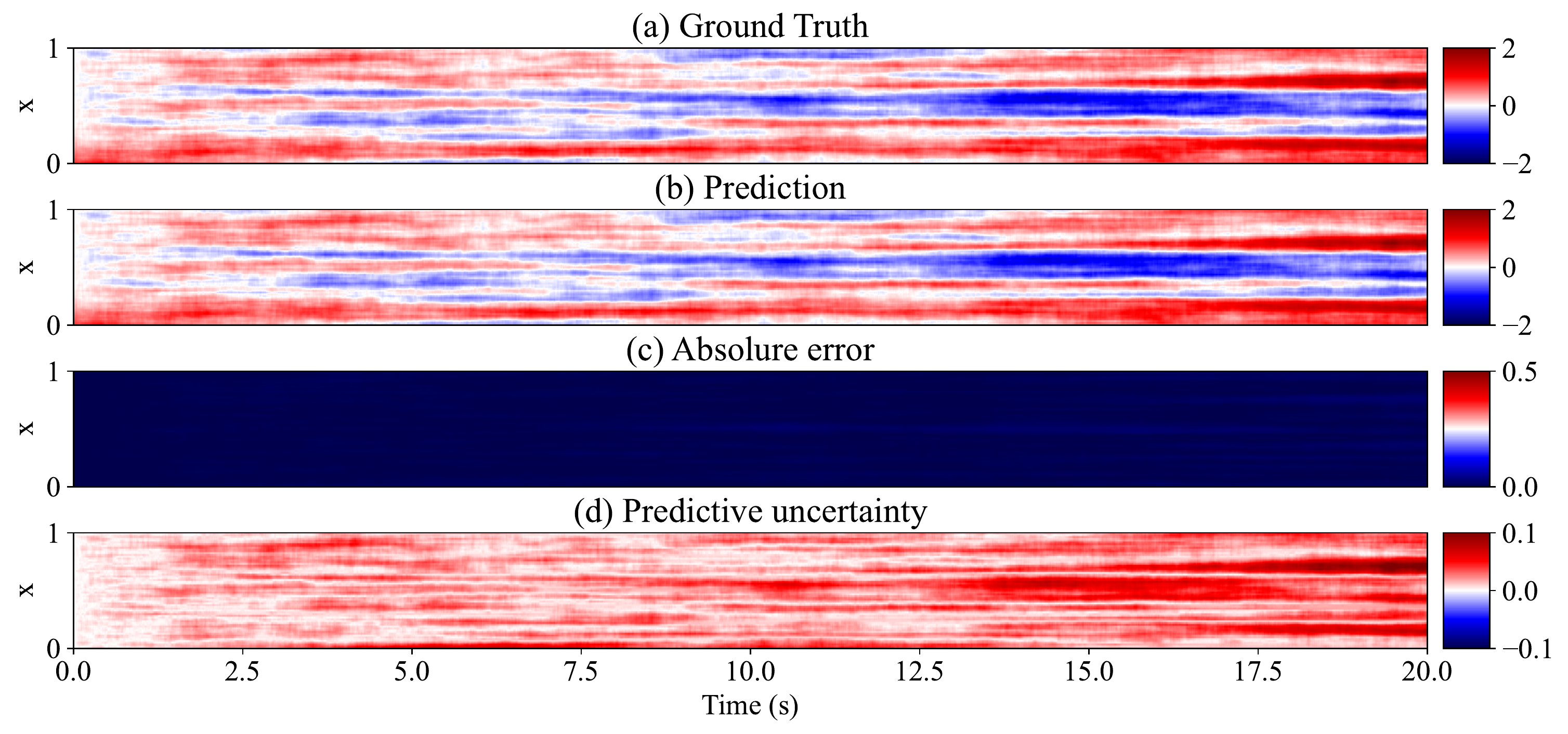}
    \caption{\textbf{Predictive performance and predictive uncertainty of the discovered 1D stochastic Heat PDE against the Ground truth}. (a) The true solution of the actual SPDE. (b) The solution of the discovered stochastic Heat PDE. (c) The absolute error between the solutions of the true and identified system. (d) Predictive uncertainty in the discovered SPDE, obtained from 200 ensembles of the solution.}
    \label{fig:heat_all}
\end{figure}
The predictive performance of the identified model for the stochastic heat equation is illustrated in Fig. \ref{fig:heat_all}. The predictions and predictive uncertainties are obtained from 200 ensembles of system solutions. Upon comparing the solution of the identified system with ground truth, it can be safely stated that the discovered model has correctly captured the stochastic evolution of the stochastic heat equation. The predictive uncertainty result is no different from the previous examples, where we can see that the uncertainty increases with time. Overall the predictive uncertainty is very small, which in combination with the prediction error indicates the accuracy of the identified SPDE model.


\subsection{Performance of the proposed VB framework}
In the proposed VB algorithm, the vector $\bm{w}^{(o)}$ (i.e., the vector of inclusion probabilities of the basis functions) was initialized using (modified) stochastic SINDy and hence, it becomes important to differentiate the accuracy of the proposed approach from the stochastic SINDy. Also, it is required to establish the efficiency of the variational framework in the proposed algorithm against Gibbs sampler-based Bayesian inference techniques in terms of both computational time and accuracy. 
The performance is assessed using three distinct metrics. The first metric is the $L^2$ error, which quantifies the disparity between the identified parameters obtained through the VB, Gibbs, and SINDy in relation to the true parameters. The corresponding results are presented in Table \ref{tab:accuracy}.
The second metric is the false positivity rate (FPR), which involves detecting false positive entries. These false positive entries refer to basis functions that are present in the identified equation but absent in the actual equation. To compute the FPR value, we employed the following formula,
\begin{equation}
    \text{FPR}\; (\%) = 100 \times \frac{\text{False positive basis functions of $\mathbf{L}$}}{\text{Total number of basis functions in the dictionary $\mathbf{L}$}}
\end{equation}
\begin{table}[ht!]
    \centering
    \caption{Performance comparison between different algorithms in terms of $L^2$ identification error, False Positive Rate, and computational time. e-SINDy represents the extended stochastic SINDy algorithm, Gibbs represents the Bayesian regression with Gibbs sampler, and VB refers to the proposed variational Bayes framework. The bold numbers indicate the best-performing metric.}
    \small
    \begin{tabular}{llllllllll}
    \toprule
    \multirow{2}{*}{Example} & \multicolumn{3}{c}{$L^2$ Error} & \multicolumn{3}{c}{False Positive Rate (\%)} & \multicolumn{3}{c}{Computational time (min)}\\
    \cmidrule(r){2-4} \cmidrule(r){5-7} \cmidrule(r){8-10}
    & e-SINDy & Gibbs & VB & e-SINDy & Gibbs & VB & e-SINDy & Gibbs & VB \\
    \midrule
    Stochastic Allen-Cahn & 0.0652 & 0.1444 & \textbf{0.0498} & \textbf{0} & \textbf{0} &\textbf{0} & \textbf{0.0031} & 656.7616 & 0.0297 \\
    \midrule
    Stochastic Nagumo & 1.5997 & 0.1482 & \textbf{0.1393} & 11.4286 & \textbf{0} &\textbf{0} & \textbf{0.0040} & 805.3746 & 0.0254 \\
    \midrule
    Stochastic Heat & 1.1586 & 0.0634 & \textbf{0.0587} & 7.1428 & \textbf{0} &\textbf{0} & \textbf{0.0125} & 676.6427 & 0.0298 \\
    \bottomrule
    \end{tabular}
    \label{tab:accuracy}
\end{table}
The third metric is computational time, which measures the efficiency of the underlying algorithm. In this comparison, the number of Gibbs iterations is fixed at 100 with a burn-in size of 50 for both drift and diffusion identification. 
The $L^2$ error values in Table \ref{tab:accuracy} clearly illustrates that the VB approach yields the smallest error in the model identification, while the identification errors in the Gibbs inference are close to the results of VB. The stochastic SINDy performs well for the stochastic Allen-Cahn equation; however, it performs poorly for other problems.
In terms of FPRs, both the VB and Gibbs inference yield a zero FPR for all three problems, while SINDy exhibits a notable FPR value other than the stochastic Allen-Cahn equation, indicating its inclusion of false basis functions. In terms of computational time, the stochastic SINDy takes the least time for model discovery, and Gibbs inference takes the maximum time. The proposed approach is significantly faster as compared to Gibbs sampling. 
To note a few points, although the stochastic SINDy is faster, the models discovered by stochastic SINDy are not always as accurate as those obtained using the proposed VB and Gibbs inference-based approaches. By increasing the number of iterations in Gibbs inference, one can achieve a higher accuracy than the presented results; however, the computational demand posed by Gibbs inference is extremely high. 
Therefore, it is safe to state that the proposed VB framework for identifying SPDE provides the right balance between accuracy and efficiency; this is particularly valuable in practical applications where dataset are large, higher-order accuracy is needed, and real-time analysis is involved. 


\section{Conclusions}\label{conclusion}
In this paper, we have proposed a novel variational Bayes framework for discovering SPDEs from data. Our method integrates the concepts of stochastic process, Bayesian statistics, and sparse learning to accurately identify the underlying SPDEs and promote interpretability in the discovered models.
By leveraging the extended Kramers-Moyal expansion, we extract the relevant target vectors and utilize sparse linear regression with SS priors to efficiently discover the underlying partial differential equations.
The proposed approach has been applied to three canonical SPDEs: (a) stochastic heat equation, (b) stochastic Allen-Cahn equation, and (c) stochastic Nagumo equation.
Our proposed method successfully identifies the exact form of the SPDE models, showcasing its capability to capture the complex interplay between deterministic parameters and random fluctuations.
The utilization of variational Bayes in our approach contributes to its computational efficiency. By avoiding computationally expensive methods like Gibbs sampling, our method provides comparable accuracy while significantly reducing the computational burden. 

While the proposed approach has demonstrated promising results, there are some limitations to consider. One major limitation is the selection of basis functions in the library. If the accurate set of basis functions is not present in the library, the model might select relevant correlated terms instead, leading to a larger equation that is less interpretable. To mitigate this, it is important to include as many basis functions as possible in the library, encompassing all potential terms that might be present in the equation representing the data.

Importantly, the proposed approach represents a significant advancement in the field of equation discovery, as it is the first attempt at discovering SPDEs from data. This breakthrough has significant implications for various scientific applications, such as climate modeling, financial forecasting, and chemical kinetics. By accurately identifying the underlying SPDEs, our method enables the development of robust mathematical models that can capture the inherent uncertainties and random influences present in natural and engineered systems. This, in turn, enhances predictive capabilities, facilitates risk assessment, and aids in decision-making processes.
Overall, we have presented an efficient and accurate method for discovering SPDEs from data. Our approach successfully identifies the underlying drift and diffusion terms of the SPDEs, allowing for a comprehensive representation of real-world phenomena. The proposed method not only provides accurate predictions but also promotes model interpretability, making it a valuable tool for scientific research and engineering applications. The discovery of SPDEs from data holds great potential for advancing our understanding of complex systems just from experimental data and improving decision-making processes in various scientific and engineering domains.

\section*{Acknowledgements} 
T. Tripura acknowledges the financial support received from the Ministry of Education (MoE), India, in the form of the Prime Minister's Research Fellowship (PMRF). S. Chakraborty acknowledges the financial support received from Science and Engineering Research Board (SERB) via grant no. SRG/2021/000467, Ministry of Port and Shipping via letter no. ST-14011/74/MT (356529), and seed grant received from IIT Delhi. R. Nayek acknowledges the financial support received from Science and Engineering Research Board (SERB) via grant no. SRG/2022/001410, Ministry of Port and Shipping via letter no. ST-14011/74/MT (356529), and seed grant received from IIT Delhi.

\section*{Declarations}


\subsection*{Conflicts of interest} The authors declare that they have no conflict of interest.


\subsection*{Code availability} Upon acceptance, all the source codes to reproduce the results in this study will be made available to the public on GitHub by the corresponding author.


\end{document}